\documentclass{article}

    \PassOptionsToPackage{numbers, compress}{natbib}

    \usepackage[final]{neurips_2022}

\usepackage[utf8]{inputenc} 
\usepackage[T1]{fontenc}    
\usepackage{hyperref}       
\usepackage{url}            
\usepackage{booktabs}       
\usepackage{amsfonts}       
\usepackage{nicefrac}       
\usepackage{microtype}      
\usepackage[dvipsnames]{xcolor}         
\usepackage{amsmath}        
\usepackage{enumitem}
\usepackage{graphicx}
\usepackage[font=small]{caption}
\usepackage{subcaption}
\usepackage{floatrow}
\usepackage{multirow}
\usepackage{siunitx}
\usepackage{mfirstuc}
 \newcommand{\ssc}[1]{\textsc{\MakeLowercase{#1}}}

\usepackage{enumitem}
\setlist{leftmargin=4.5mm}

\newcommand\numberthis{\addtocounter{equation}{1}\tag{\theequation}}

\title{Additive MIL: Intrinsically Interpretable Multiple Instance Learning for Pathology}

\author{%
  Syed Ashar Javed \\
  PathAI Inc \\
  \texttt{ashar.javed@pathai.com} \\
   \And
  Dinkar Juyal \\
  PathAI Inc \\
  \texttt{dinkar.juyal@pathai.com} \\
  \And
  Harshith Padigela \\
  PathAI Inc \\
  \texttt{harshith.padigela@pathai.com} \\
  \And
  Amaro Taylor-Weiner \\
  PathAI Inc \\
  \texttt{amaro.taylor@pathai.com}
  \And
  Limin Yu \\
  PathAI Inc \\
  \texttt{limin.yu@pathai.com}
  \And
  Aaditya Prakash \\
  PathAI Inc \\
  \texttt{adi.prakash.ml@gmail.com}
}

\begin{document}
\maketitle

\begin{abstract}
Multiple Instance Learning (\ssc{MIL}) has been widely applied in pathology towards solving critical problems such as automating cancer diagnosis and grading, predicting patient prognosis, and therapy response. Deploying these models in a clinical setting requires careful inspection of these black boxes during development and deployment to identify failures and maintain physician trust. In this work, we propose a simple formulation of \ssc{MIL} models, which enables interpretability while maintaining similar predictive performance. Our Additive \ssc{MIL} models enable spatial credit assignment such that the contribution of each region in the image can be exactly computed and visualized. We show that our spatial credit assignment coincides with regions used by pathologists during diagnosis and improves upon classical attention heatmaps from attention \ssc{MIL} models. We show that any existing \ssc{MIL} model can be made additive with a simple change in function composition. We also show how these models can debug model failures, identify spurious features, and highlight class-wise regions of interest, enabling their use in high-stakes environments such as clinical decision-making.
\end{abstract}

\section{Introduction}

Histopathology is the study and diagnosis of disease by microscopic inspection of tissue. Histologic examination of tissue samples plays a key role in both clinical diagnosis and drug development. It is regarded as medicine's ground truth for various diseases and is important in evaluating disease severity, measuring treatment effects, and biomarker scoring \citep{walk2009role}. A differentiating feature of digitized tissue slides or whole slide images (\ssc{WSI}) is their extremely large size, often billions of pixels per image. In addition to being large, \ssc{WSI}s are extremely information dense, with each image containing thousands of cells and detailed tissue regions that make manual analysis of these images challenging. This information richness makes pathology an excellent application for machine learning, and indeed there has been tremendous progress in recent years in applying machine learning to pathology data \citep{wang2016deep, campanella2019clinical, kather2019deep, wulczyn2020deep, diao2021human, bosch2021machine, glass2021machine, echle2021deep}.

The most important applications of ML in digital pathology involve predicting patient's clinical characteristics from a \ssc{WSI} image. Models need to be able to make predictions about the entire slide involving all the patient tissue available; we refer to these predictions as "slide-level". To overcome the challenges presented by the size of these images, previous methods have used smaller hand engineered representations, built from biological primitives in tissue such as cellular composition and structures \citep{diao2021human}. Another common way to overcome the challenges presented by the size of \ssc{WSI}s is to break the slide into thousands of small patches, train a model with these patches to predict the slide-label, and then use a secondary model to learn an aggregation function from patch representations to slide-level label \citep{campanella2019clinical}. Both methods are not trained in an end-to-end manner and suffer from sub-optimal performance. The second method also suffers from an incorrect assumption that each patch from a slide has the same label as the overall slide \citep{ilse2018attention}.

Multiple Instance Learning \citep{maron1997framework} is a weakly supervised learning technique which attempts to learn a mapping from a set of instances (called a bag) to a single label associated with the whole bag. \ssc{MIL} can be applied to pathology by treating patches from slides as instances which form a bag and a slide-level label is associated with each bag to learn a bag predictor. This circumvents the need to collect patch-level labels and allows end-to-end training from a \ssc{WSI}. The \ssc{MIL} assumption that at least one patch among the set of patches is associated with the target label works well for many biological problems. For example, the \ssc{MIL} assumption holds for the task of cancer diagnosis; a sufficiently large bag of instances or patches from a cancerous slide will contain at least one cancerous patch whereas a benign slide will never contain a cancerous patch. 
In recent years, attention based pooling of patches has been shown to be successful for \ssc{MIL} problems \citep{ilse2018attention}. Using neural networks with attention \ssc{MIL} has become the standard for end-to-end pathology models as it provides a powerful, yet efficient gradient based method to learn a slide-to-label mapping. In addition to superior performance, these models encode some level of spatial interpretability within the model through visualization of highly attended regions. 

The sensitive nature of the medical imaging domain requires deployed machine learning models to be interpretable for multiple reasons. First, it is critical that models do not learn spurious shortcuts over true signal \citep{degrave2021ai, schmitt2021hidden} and can be debugged if such failure modes exist. Interpretability and explainability methods have been shown to help identify some of these data and model deficiencies \citep{hagele2020resolving, lu2021data, anders2022finding, tang2021data}. Secondly, for algorithms in medical decision-making, accountability and rigorous validation precedes adoption \citep{london2019artificial, javed2022rethinking}. Interpretable models can be easier to validate and thus build trust. Specifically, users can verify that model predictions are generated using biologically concordant features that are supported by scientific evidence and are similar to the those identified by human experts. Thirdly, use-cases involving a human expert such as decision-support require the algorithm to give a visual cue which highlights the regions to be examined more carefully. In these applications, a predicted score is insufficient and needs to be complemented with a highlighted visual region associated with the model's prediction \citep{bulten2021artificial}.

For ML models in pathology, spatial credit assignment can be defined as attributing model predictions to specific spatial regions in the slide. Various post-hoc interpretability techniques like gradient based methods \citep{pirovano2020improving} and  Local Interpretable Model-agnostic Explanation (\ssc{LIME}) \citep{wang2019machine} have been used to this end. However, gradient based methods which try to construct model-dependent saliency maps are often insensitive to the model or the data \citep{kindermans2019reliability, adebayo2018sanity, atrey2019exploratory}. This makes these post-hoc methods unreliable for spatial attribution as they provide poor localization and do not reflect the model's predictions \citep{hagele2020resolving}. Model-agnostic methods like Shapley values or LIME involve intractable computations for large image data and thus need approximations like locally fitting explanations to model predictions, which can lead to incorrect attribution \citep{molnar2022general}. Applying Attention \ssc{MIL} \citep{ilse2018attention} in weakly supervised problems in pathology leads to learning of the attention scores for each patch. These scores can be used as a proxy for patch importance, thus helping in spatial credit assignment. This way of interpreting \ssc{MIL} models has been used commonly in the literature to create spatial heatmaps, image overlays that indicate credit assignment, for free without applying any post-hoc technique \citep{yao2020whole, lu2021data, li2021multi, shao2021transmil, li2021dual}. The attention values that scale patch feature representations have a non-linear relationship to the final prediction, making their visual interpretation inexact and incomplete. We discuss these issues in detail in the next section.

To resolve these issues, we propose a simple formulation of \ssc{MIL} which induces intrinsically interpretable heatmaps. We refer to this model as Additive \ssc{MIL}. It  allows for exact decomposition of a model prediction in  terms of spatial regions of the input. These models are inspired by Generalized Additive Models (\ssc{GAM}s) \citep{hastie2017generalized} and Neural Additive Models (\ssc{NAM}s) \citep{agarwal2021neural}, but instead of being applied to arbitrary features, they are grounded as patch instances in the \ssc{MIL} formulation which allows exact credit assignment for each patch in a bag. Specifically, we achieve this by constraining the space of \textit{predictor functions} (the classification or regression head at the final layer) in the \ssc{MIL} setup to be additive in terms of instances. Therefore, the exact contribution of each patch or instance in a bag can be traced back from the final predictions. We show that these additive scores reflect the true marginal contribution of each patch to a prediction and can be visualized as a heatmap on a slide for various applications like model debugging, validating model performance, and identifying spurious features. We also show that these benefits are achieved without any loss of predictive performance even though the predictor function is now fixed to be additive. This is critical as the accuracy-interpretability trade-off has been an active area of research and has deep implications for applications in medical imaging. Trading off performance for interpretability might make sense for improving validation and in turn adoption of ML tools, however it raises ethical questions about deploying sub-optimal models \citep{rudin2019stop, linardatos2020explainable, caruana2020intelligible}. Furthermore, since our work is orthogonal to previous advancements in \ssc{MIL} modeling, we  show that previous methods can be made additive by a simple switching of function composition at the last layer of the model, making it applicable to all \ssc{MIL} models where instance-level credit assignment is important.





\section{MIL Models \& Interpretability}
\subsection{Interpretability in Attention MIL}
\label{section_attn_mil_limitations}
An attention \ssc{MIL} model \citep{ilse2018attention} can be seen as a 3-part model consisting of a featurizer ($f$) , typically a deep CNN, an attention module ($m$) which induces a soft attention over the $N$ patches and is used to scale each patch feature, and a predictor ($p$) which takes the attended patch representations, aggregates them using a permutation invariant function like sum pooling over the $N$ patches, and then outputs a prediction. This \ssc{MIL} model $g(x)$ is given as:
\begin{align*}
    g(x) &= (p \circ m \circ f)(x) \numberthis  \label{eqn_attention_mil_model_1} \\
    m_i(x) &= \alpha_i f(x_i) \quad\text{where}\quad \alpha_i = softmax_i(\psi_m(x))  \numberthis  \label{eqn_attention_mil_model_2} \\
    p(x) &= \psi_p(\sum_{i=1}^{N} m_i(x)) \numberthis  \label{eqn_attention_mil_model_3} 
\end{align*}
where $\psi_m$ and $\psi_p$ are MLPs with non-linear activation functions.

The attention scores $\alpha_i$ learned by the model can be treated as patch importance scores and have been used previously to interpret \ssc{MIL} models \citep{yao2020whole, lu2021data, shao2021transmil, li2021dual}. However, there are four issues in doing spatial attribution using these attention scores. For example, consider the task of classifying a slide into benign, suspicious or malignant:
\begin{enumerate}[label=(\alph*)]
\item Since the attention weights are used to scale the patch features used for the prediction task, a high attention weight only means that the patch might be needed for the prediction downstream. Therefore, a high attention score for a patch can be a necessary but not sufficient condition for attributing a prediction to that patch. Similarly, patches with low attention can be important for the downstream prediction since the attention scores are related non-linearly to the final classification or regression layer. For example, in a malignant slide, non-tumor regions might get highlighted by the attention scores since they need to be represented at the final classification layer to provide discriminative signal. However, this does not imply malignant prediction should be attributed to non-malignant regions, nor that these regions would be useful to guide a human expert.
\item The patch's contribution to the final prediction can be either positive (excitatory) or negative (inhibitory), however attention scores do not distinguish between the two. A patch might be providing strong negative evidence for a class but will be highlighted in the same way as a positive patch. For example, benign mimics of cancer are regions which visually look like cancer, but are normal benign tissue \citep{trpkov2018benign}. These regions are useful for the model to provide negative evidence for the presence of cancer and thus might have high attention scores. While attending to these regions may be useful to the model, they may complicate human interpretation of resulting heatmaps.
\item Attention scores do not provide any information about the class-wise importance of a patch, but only that a patch was weighted by a certain magnitude for generating the prediction. In the case of multiclass classification, this becomes problematic as a high attention scores on a patch can mean that it might be useful for any of the multiple classes. Different regions in the slide might be contributing to different classes which are indistinguishable in an attention heatmap. For example, if a patch has high attention weight for benign-suspicious-malignant classification, it can be interpreted as being important for any one or more of the classes. This makes the attention scores ineffective for verifying the role of individual patches for a slide-level prediction.
\item Using attention scores to assess patch contribution ignores patch interactions at the classification stage. For example, two different tumor patches might have moderate attention scores, but when taken together in the final classification layer, they might jointly provide strong and sufficient information for the slide being malignant. Thus, computing marginal patch contributions for a bag needs to be done at the classification layer and not the attention layer since attention scores do not capture patch interactions and thus can under or over estimate contributions to the final prediction.
\end{enumerate}

These difficulties in interpreting attention \ssc{MIL} heatmaps motivate the formulation of a traceable predictor function, where model predictions can be exactly specified in terms of patch contributions (both positive and negative) for each class.
\begin{figure}[h!]
\centering
\includegraphics[scale=0.55]{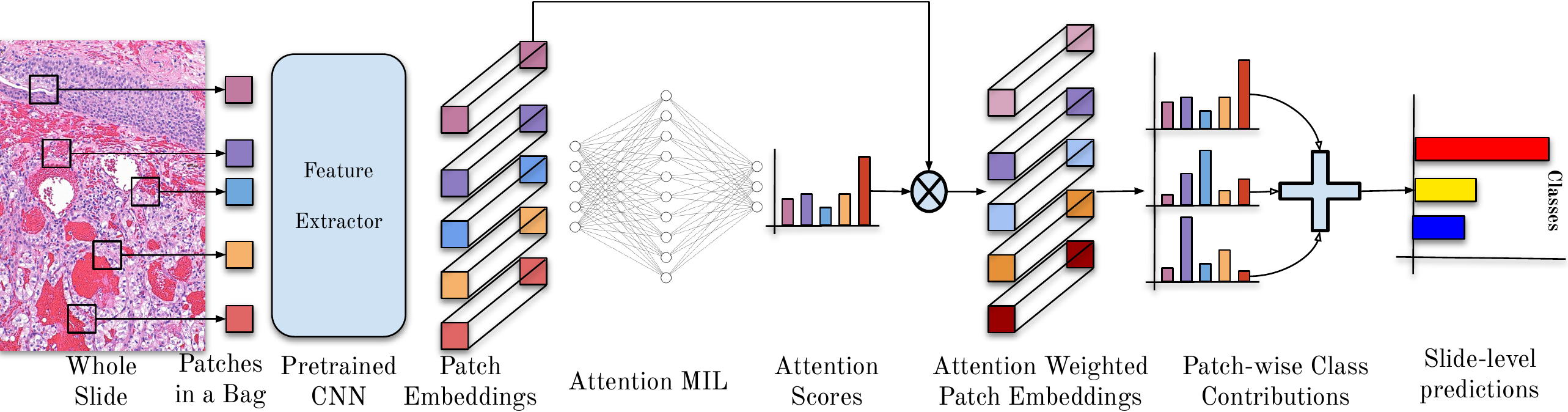}
\caption{Additive MIL Model}
\label{additive_mil}
\end{figure}

\subsection{Additive MIL Models}
We first define the desiderata for a visual interpretability method for \ssc{MIL} models:
\begin{enumerate}
    \item The method should be intrinsic to the model and not be a post-hoc method. This prevents incorrect assumptions about the model and does not require post-hoc modeling. It also prevents many pitfalls of traditional saliency methods \citep{kindermans2019reliability, adebayo2018sanity}.
    \item Attribution in the \ssc{MIL} setup should be in terms of instances only. For pathology, this means that the prediction should be attributed to individual patches. This constraint enables expression of bag predictions in terms of marginal instance contributions.
    \item Should reflect model's sensitivity and invariances by reliably mirroring its functioning \citep{leavitt2020towards}.
    \item Should distinguish between excitatory and inhibitory patch contributions. Should also provide per-class contributions for classification problems.
\end{enumerate}

To enable the desired instance-level credit assignment in \ssc{MIL}, we re-frame the final predictor to be an additive function of individual instances. This translates to a simple switching of the function composition in Equation ~\ref{eqn_attention_mil_model_3}:
\begin{align*}
    p_{Additive}(x) &= \sum_{i=1}^{N} \psi_p(m_i(x)) \numberthis  \label{eqn_additive_mil_model_predictor} 
\end{align*}

Making this change results in the final predictor only being able to implement patch-additive functions on top of arbitrarily complex patch representations. This provides both complexity of the learned representations as well as a traceable patch contribution for a given prediction which solves the spatial credit assignment problem. $\psi_p(m_i(x))$ is the class-wise contribution for patch $i$ in the bag. At inference, $\psi$ produces a $R^{C \times N}$ for a classification problem where $C$ is the number of classes and $N$ is the number of patches in a bag. Thus, we get a class-wise score for each patch, which when summed gives the final logits for the prediction problem. These scores can be visualized by constructing a heatmap from the visual representation of patch-wise contributions for each class. The sign of the patch contribution decides whether the patch is excitatory or inhibitory towards each class since positive values add to the final logit while negative values bring down the final class logit. In the next section, we prove that the instance contribution obtained from an Additive \ssc{MIL} model is exactly equivalent to the actual marginal contribution of that patch to a model's prediction.

\subsection{Proof of equivalence between Additive MIL and Shapley Values}
We highlight the spatial credit assignment properties of an Additive \ssc{MIL} model by proving its equivalence to Shapley values \citep{shapley1953value}. Shapley value is game theoretic concept used for calculating the optimal marginal contribution of each player in a n-player coalition game with a given total payoff. In machine learning, it's used to interpret models predictions by decomposing them in terms of their marginal feature contributions. There have been various applications of using Shapley values for feature credit assignment in the interpretability literature \citep{lundberg2017unified, messalas2019model,tang2021data, sundararajan2020many,aas2021explaining}. However for most practical models, the computational complexity grows exponentially, thus requiring approximations \citep{van2022tractability}. Here, we show that our additive MIL formulation is equivalent to Shapley values without any approximation.

The Additive \ssc{MIL} model, $g$ is defined for a \ssc{MIL} problem with instances $x_i$ as input:
\begin{align*}
    g(x) = \sum_{n=1}^{N} \psi_p(\alpha_i f(x_i)) \numberthis  \label{eqn_additive_mil_model}
\end{align*}
where $\alpha_i$ is the attention weight for the $i^{th}$ instance, $f$ is the function encoding each instance into a feature representation, and $\psi_p$ is predictor function which maps instance representation to the model output (e.g. logits for classification models).

\textbf{Theorem 1:} \textit{The marginal instance contribution from an Additive \ssc{MIL} model, $g(x_i)$ is proportional to the Shapley value of that instance, $\phi_i$.}
\begin{align*}
    g(x_i) \propto \phi_i(V, x) = \sum_{S\subseteq F \backslash i} \frac{|S!| (|F|-|S|-1|!)}{|F!|} V_{S \cup i}(x_{S \cup i}) - V_{S}(x_{S}) \numberthis \label{eqn_additive_shapley}
\end{align*}
\textit{Consequence:} Additive \ssc{MIL} scores ensure optimal credit assignment across instances of an \ssc{MIL} bag. Thus each bag-level prediction in \ssc{MIL} can be exactly decomposed into marginal instance contributions given by Additive \ssc{MIL} scores and provide model interpretability.


\textit{Proof:} 
The proof is in Appendix \ref{shapley_proof}

\subsection{Features of Additive MIL Models}
\textbf{Exact marginal patch contribution towards a prediction}.  Additive \ssc{MIL} models provide exact patch contribution scores which are additively related to the prediction. This additive coupling of the model and the interpretability method makes the spatial scores precisely mirror the invariances and the sensitivities of the model, thus making them intrinsically interpretable.

\textbf{Class-wise contributions}. Additive \ssc{MIL} models allow decomposing the patch contributions and attributing them to individual classes in a classification problem. This allows us to not just assign the prediction to a region, but to also see which class it contributes to specifically. This is helpful in cases where signal for multiple classes exist within the same slide. 

\textbf{Distinction between excitatory and inhibitory contributions}. Additive \ssc{MIL} models allow for both positive and negative contributions from a patch. This can help distinguish between areas which are important because they provide evidence for the prediction and those which provide evidence against.

\section{Experiments \& Results}

\subsection{Experimental Setup \& Datasets}
We perform various experiments to show the benefits of using Additive \ssc{MIL} models for interpretability in pathology problems. Concretely, we show the following results:

\begin{itemize}
    \item Additive \ssc{MIL} models provide intrinsic spatial interpretability without any loss of predictive performance as compared to more expressive, non-additive models.
    \item Any pooling-based \ssc{MIL} model can be made additive by reformulating the predictor function and leads to predictive results similar to the original model.
    \item Additive \ssc{MIL} heatmaps yield better alignment with region-annotations from an expert pathologist than Attention \ssc{MIL} heatmaps.
    \item Additive \ssc{MIL} heatmaps provide more granular information like class-wise spatial assignment and excitatory and inhibitory patches which is missing in attention heatmaps. This can be useful for applications like model debugging.
\end{itemize}

We consider $3$ different datasets and $2$ different problems for our experiments. The first problem is the prediction of cancer subtypes in non-small cell lung carcinoma (\ssc{NSCLC}) and renal cell carcinoma (\textsc{\ssc{RCC}}), both of which use the \ssc{TCGA} dataset \citep{weinstein2013cancer}. The second problem is the detection of metastasis in breast cancer using the Camelyon16 dataset \citep{bejnordi2017diagnostic}. \ssc{TCGA} \ssc{RCC} contains $966$ whole slide images (\ssc{WSI}s) with three histologic subtypes - \ssc{KICH} (chromophobe \ssc{RCC}), \ssc{KIRC} (clear cell \ssc{RCC}) and \ssc{KIRP} (papillary \ssc{RCC}). We extract $768k$ patches from this dataset which translates to an average of $795$ patches per slide and $16k$ total bags. \ssc{TCGA} \ssc{NSCLC} has $1002$ \ssc{WSI}s, with $538$ slides belonging to subtype \ssc{LUAD} (Lung Adenocarcinoma) and $464$ to \ssc{LUSC} (Lung Squamous Cell Carcinoma). We extract $1.465$ million patches from this dataset which translates to an average of $1462$ patches per slide and $30.5k$ total bags. Camelyon16 contains $267$ \ssc{WSI}s for training and $129$ for testing with a total of $159$ malignant slides and $237$ benign slides. We extract $510k$ patches from this dataset which translates to an average of $1286$ patches per slide and $10.6k$ total bags. These numbers point to the diversity in the dataset size in terms of number of slides, number of bags, and the label imbalance.

\subsection{Implementation Details}
Both \ssc{TCGA} datasets were split into 60/15/25 (train/val/test) as done previously \citep{shao2021transmil} while ensuring no data leakage at a case level. For Camelyon16, we use the original splits provided with the dataset.
For training the models, a bag size of $48$-$1600$ patches and batch size of $16$-$64$ was experimented with and the best one chosen using cross-validation. The patches were sampled from non-background regions for all datasets at a resolution of 1 microns per pixel without any overlap between adjacent patches. An ImageNet pre-trained Shufflenet \citep{ma2018shufflenet} was used as the feature extractor and the entire model was trained with \ssc{ADAM} optimizer \citep{kingma2014adam} and a learning rate of $1$e-$4$.
For inference, multiple bag predictions were aggregated using a majority vote to get the final slide-level prediction. \ssc{AUROC} (area under the receiver operating curve) scores were generated using the proportion of bags predicting the majority label as the class assignment probability. For \ssc{TCGA}-\ssc{RCC}, we compute macro average of 1-vs-rest \ssc{AUROC} across the 3 classes.
The attention scores were obtained by directly taking the raw outputs for each patch from the attention module. For additive patch contributions, the patch-wise class contributions were taken and converted to a bounded patch contribution value using a sigmoid function. This yielded excitatory scores in the range of $0.5-1$ and inhibitory scores in the range of $0-0.5$. Both the attention and additive patch-wise scores were used for generating a heatmap as an overlay on the slide with Attention \ssc{MIL} heatmaps having a single value per patch and Additive \ssc{MIL} heatmaps having $C$ values per patch where $C$ is the number of classes. All training and inference runs were done on Quadro \ssc{RTX} $8000$, and it takes $3$ to $4$ hours to train the model with four \ssc{GPU}s.

\subsection{Experimental Results}

\begin{table}[]
	\caption{Comparison of predictive performance on Camelyon16, \ssc{TCGA} \ssc{NSCLC} \& \ssc{RCC} datasets.}
	\label{performance_table}
	\centering 
	\begin{tabular}{@{}c||cc||cc|cc@{}}
		\toprule
		\multirow{2}{*}{\textbf{Method}}
		&\multicolumn{2}{c}{\textbf{Camelyon16}}
		&\multicolumn{2}{c}{\textbf{\ssc{TCGA} \ssc{NSCLC}}}     & \multicolumn{2}{c}{\textbf{\ssc{TCGA} \ssc{RCC}}}        \\ 
		\cmidrule(l){2-7} 
		& Accuracy        & AUC
		& Accuracy        & AUC             & Accuracy         & AUC              \\ 
		\midrule
		\midrule
		Mean Pooling \ssc{MIL}              & 0.751 & 0.707  & 0.830          & 0.925          & 0.918           & 0.980           \\
		Mean Pooling \ssc{MIL} + Additive     & 0.734 & 0.687 & 0.866           & 0.924          & 0.902           & 0.974           \\
		\midrule
		Attention \ssc{MIL} \textsc{[\ssc{ABMIL}]} \cite{ilse2018attention}              & 0.773 & 0.750  & 0.883          & 0.946          & 0.878           & 0.978           \\
		Attention \ssc{MIL} + Additive     & 0.830 & 0.846 & 0.886           & 0.941          & 0.915           & 0.983           \\
		\midrule
		Trans\ssc{MIL} \cite{shao2021transmil}            &0.805 & 0.775 & 0.878          & 0.932          & 0.915           & 0.983           \\
		Trans\ssc{MIL} + Additive            & 0.805 & 0.844 & 0.895          & 0.934          & 0.911           & 0.986           \\
		\bottomrule
	\end{tabular}
\end{table}

\subsubsection{Predictive Performance of Additive MIL Models \& Applicability to Previous Methods}
We compare Additive \ssc{MIL} models with existing techniques in terms of predictive performance on 3 different datasets. We implement a mean-pooling based \ssc{MIL} baseline without any attention, the standard Attention \ssc{MIL} model (\ssc{ABMIL}) and a transformer based \ssc{MIL} model, Trans\ssc{MIL} which is the  state-of-the-art on these three datasets (for comparison of Trans\ssc{MIL} with other methods, refer \citep{shao2021transmil}). Table \ref{performance_table} shows how Additive \ssc{MIL} models achieve comparable or superior performance to the standard Attention \ssc{MIL} model. In the case of improved performance, we hypothesize that the additive constraint regularizes the model and limits overfitting in comparison to previous approaches. This is particularly relevant to pathology datasets that often have less than one thousand slides. The results in the table also demonstrate how previous techniques like Trans\ssc{MIL} can be made additive by switching the function composition of the classifier layer as done in Equation \ref{eqn_attention_mil_model_3} and \ref{eqn_additive_mil_model_predictor}. This property is general, thus any high performing \ssc{MIL} method can be converted to an Additive \ssc{MIL} model. Implementing the additive formulation gives nearly all the benefits of modeling complexity from previous methods, while enabling spatial interpretability without any loss of predictive performance.

\begin{figure}[]
\centering
\label{segmentation_performance_camelyon}
\hspace*{-0.2in}
\includegraphics[scale=0.46]{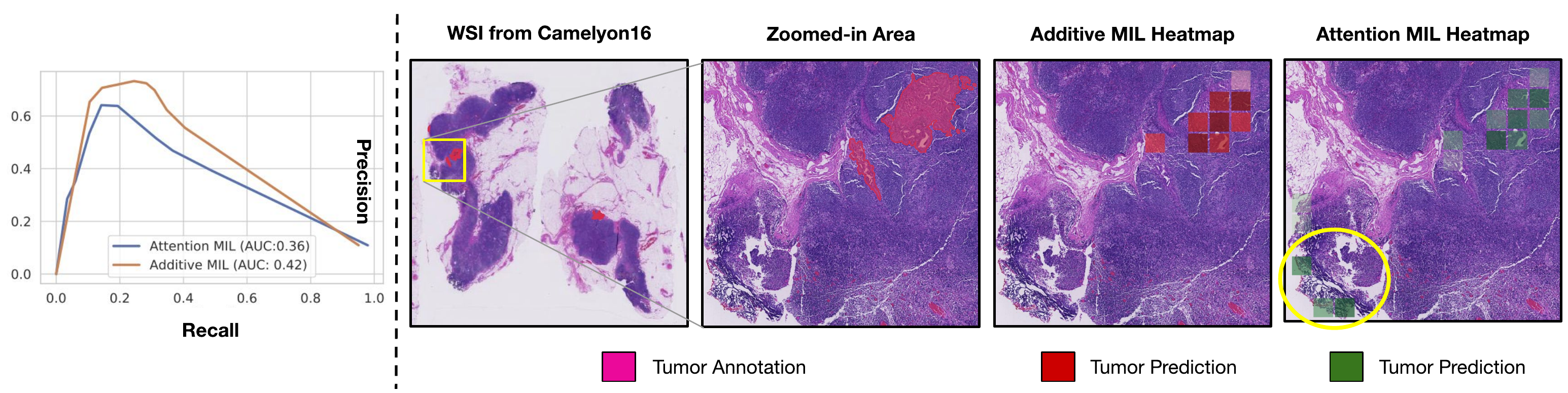}
\caption{Comparison of Additive \& Attention \ssc{MIL} heatmaps at detecting annotated cancer regions in Camelyon16. Attention \ssc{MIL} heatmaps have lower precision and detect false-positives as highlighted in the yellow circle.}
\label{heatmap_comparison_camelyon}
\end{figure}

\begin{figure}[]
\centering
\begin{subfigure}{0.9\textwidth}
  \centering
  \includegraphics[width=1\linewidth, scale=0.3]{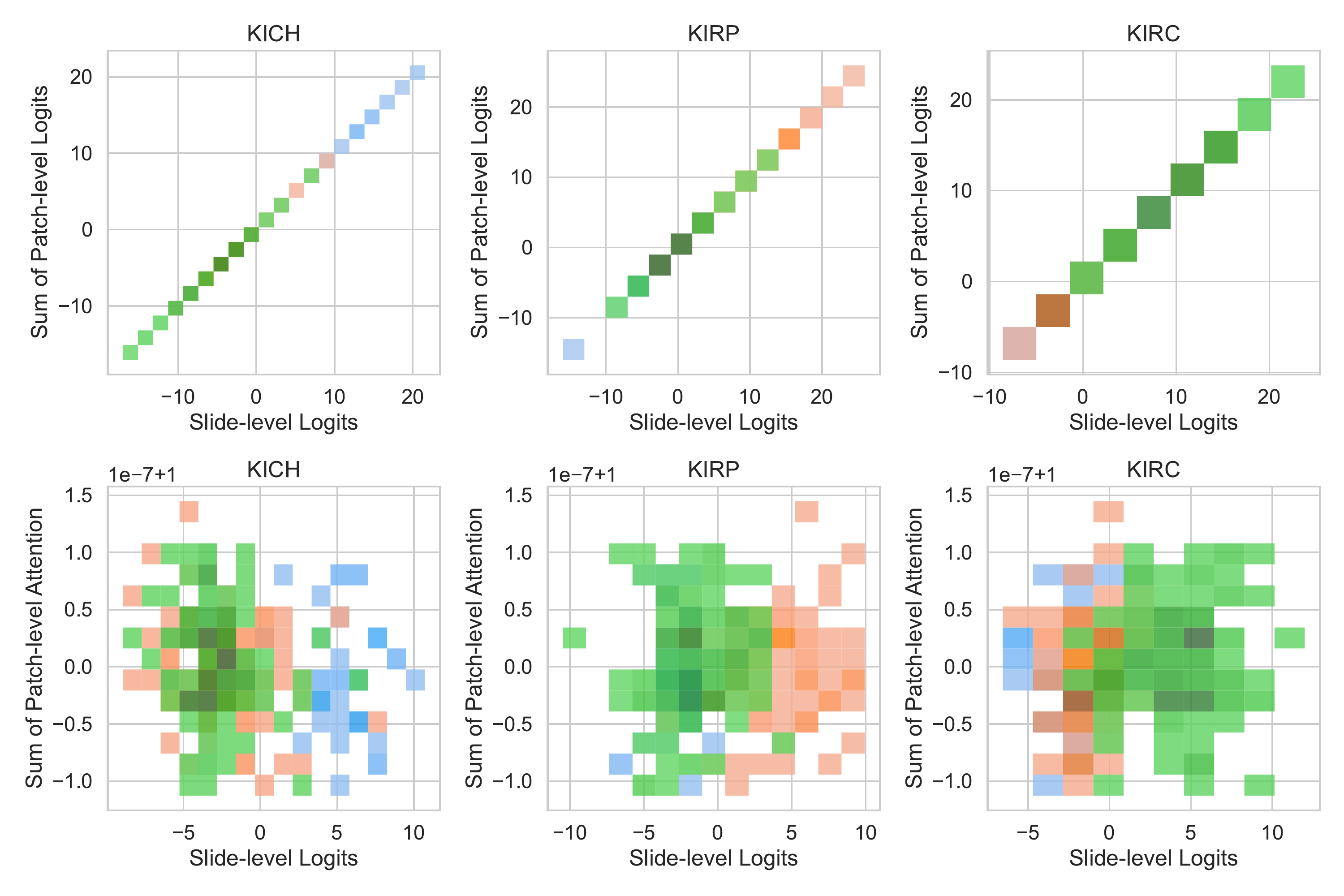}
\end{subfigure}%
\begin{subfigure}{.1\textwidth}
  \centering
  \includegraphics[width=1\linewidth]{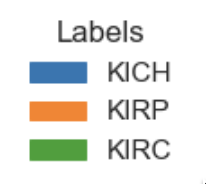}
\end{subfigure}
\hspace*{-0.2in}
\caption{Figure shows the alignment between the slide-level predicted logits and patch contributions from the Additive and the Attention models on \ssc{TCGA} \ssc{RCC}. Top-row: Y-axis shows the sum of patch contribution in a bag for \textbf{Additive}. Bottom-row: Y-axis shows the median score from top-10\% patches in a bag for \textbf{Attention}. The columns represent the slide-level logits for each class. The colors represent the ground-truth. Additive contributions are linear while Attention is not.}
\label{patch_plot}
\end{figure}

\subsubsection{Region-level Alignment of Additive Heatmaps with Expert Pathologist}
Here we compare heatmaps obtained through Additive \ssc{MIL} models with attention heatmaps and evaluate both against region-level annotations from an expert pathologist. We use the Camelyon16 dataset where the goal is to classify the slide as benign or malignant. Since the cancer foci are very localized and often occupy less than 1\% of the slide, the task of generating localized cancer heatmaps in a weakly supervised setup is very challenging. We obtained exhaustive segmentation annotations for cancer regions from a board-certified pathologist on the Camleyon16 test set. We trained an Additive \ssc{MIL} model and generated both the traditional attention heatmaps using the patch-level attention scores and Additive \ssc{MIL} heatmaps using the patch contributions. We compare the patch level precision-recall curves at different thresholds of the heatmap. Note that this comparison controls for model performance as both heatmaps are generated from the same model. Figure \ref{heatmap_comparison_camelyon} shows this comparison. At low thresholds, nearly all patches are highlighted and we see a high recall and low precision for both methods. As we increase this threshold we see higher precision and lower recall. The Additive \ssc{MIL} heatmaps (\ssc{AUPRC} $0.42$) highlighted cancer regions more precisely and sensitively than traditional attention heatmaps (\ssc{AUPRC} $0.36$), which detect more false-positives (Figure \ref{heatmap_comparison_camelyon}). If we choose the best operating point of both the curves, we find that the best F1 score for the attention heatmap is $0.43$ as compared to $0.47$ from the Additive heatmap. These experiments demonstrate the superior performance of Additive \ssc{MIL} heatmaps in localizing areas of interest.




\subsubsection{Faithful Representation of Patch-level Contributions to Slide-level Predictions}
Attention heatmaps are often used to signal regions of interest in a slide, however as argued in Section \ref{section_attn_mil_limitations}, it is not straightforward to draw conclusion regarding the importance and contribution of `attended' areas towards the model prediction. Additive \ssc{MIL} guarantees that each patch's contribution is linear and thus faithfully represents its marginal contribution toward the slide-level prediction. This property is shown in Figure~\ref{patch_plot}, where the linear relationship of Additive \ssc{MIL} (top row) is clear. In contrast, when considering the attention scores of the most attended patches (top 10\% of patches), there is no relationship with the final predictions (shown in the bottom row). 

\begin{figure}[]
\centering
\hspace*{-0.1in}
\includegraphics[scale=0.17]{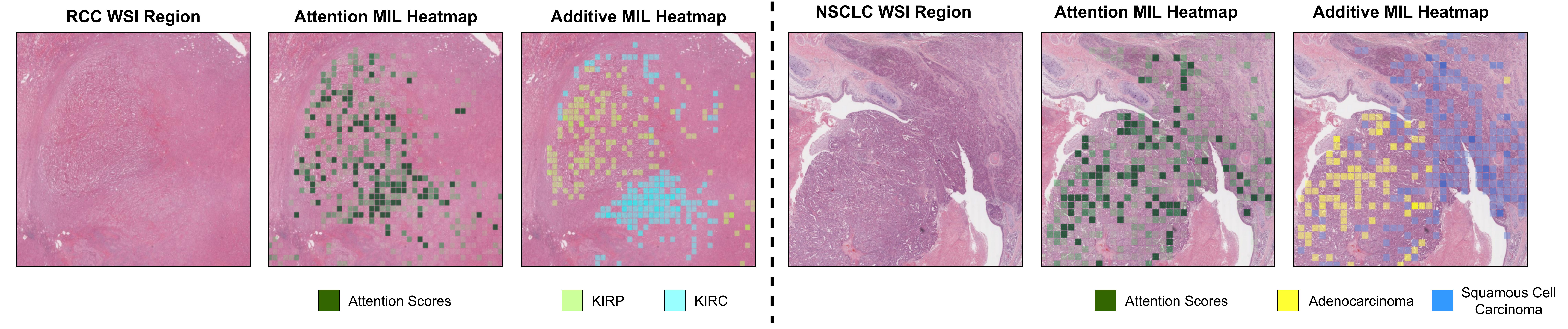}
\caption{Even in cases where attention heatmap identifies the same region as an Additive \ssc{MIL} heatmap, the ability of the latter to show granular class-wise patch contributions can lead to better interpretation. The left $3$ column (\ssc{RCC}) shows the Additive \ssc{MIL} heatmaps: \textcolor{Cyan}{\ssc{KIRC}} regions in cyan and \textcolor{LimeGreen}{\ssc{KIRP}} in green. The right $3$ columns show \textcolor{Goldenrod}{Adenocarcinoma} regions in yellow and \textcolor{Blue}{Squamous cell carcinoma} in blue. Attention heatmaps are incapable of visualizing such class-level information.}
\label{figure_multi_class_heatmaps}
\end{figure}

\subsubsection{Qualitative Assessment of Multi-Class \& Excitatory-Inhibitory Heatmaps}
\begin{figure}[]
\centering
\includegraphics[scale=0.235]{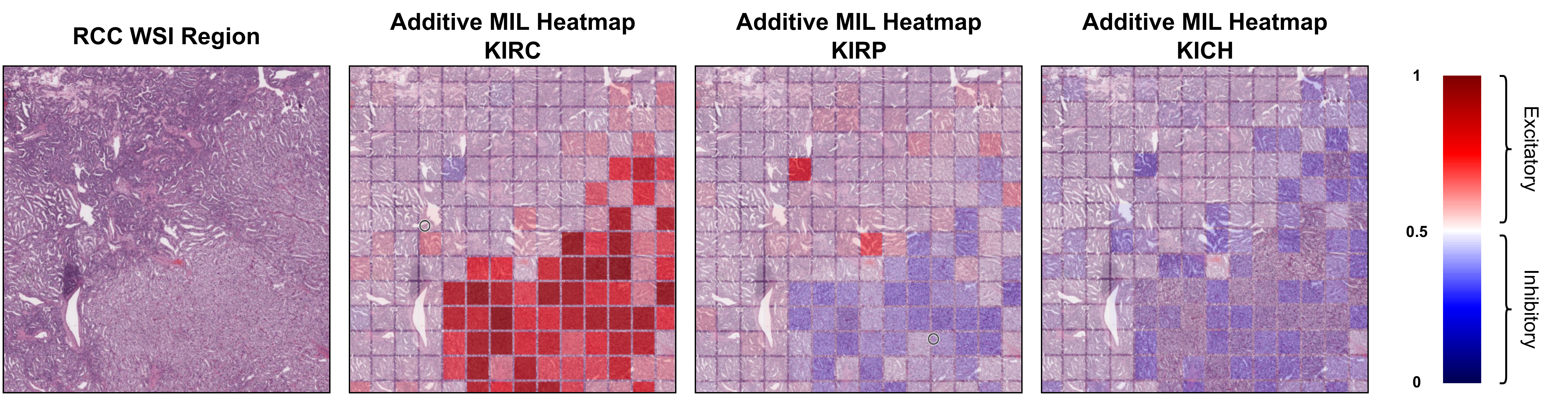}
\caption{Additive \ssc{MIL} heatmaps provide excitatory and inhibitory patch contributions for each class which can be analyzed to understand how each region is voting for or against each class. Red denotes excitatory while blue denotes inhibitory contributions for each class. In the same \ssc{RCC} slide, regions containing morphological signal for multiple subtypes can be seen and corroborated using the excitatory-inhibitory patches from different regions, thus helping in model evaluation.}
\label{figure_excite_inhibit_heatmaps}
\end{figure}

We highlight the benefits of having class-wise excitatory-inhibitory contributions for each spatial region in a slide. Figure \ref{figure_multi_class_heatmaps} shows zoomed-in regions of two slides from \ssc{TCGA} \ssc{RCC} and \ssc{NSCLC}. In these examples, the attention heatmaps do highlight tissue regions predictive of the cancer subtype but don't provide information about the association of patches to classes. In contrast, the Additive \ssc{MIL} heatmaps show precisely how each patch contributes to each class, and in turn the final prediction. Figure \ref{figure_excite_inhibit_heatmaps} shows the information about excitatory and inhibitory patches for different classes. The Additive \ssc{MIL} heatmaps for each class are visualized by the same colorbar where red denotes excitatory patches and blue denotes inhibitory ones. The \ssc{RCC} \ssc{WSI} is labeled as \ssc{KIRC}, but the selected region contains two subtypes, namely \ssc{KIRC} and small regions of \ssc{KIRP}, as evident from the raw slide. The Additive \ssc{MIL} heatmaps accurately show bottom right region being excitatory for \ssc{KIRC}, but inhibitory for the other two whereas the top left region is only excitatory for \ssc{KIRP} and inhibitory for two other two. All patches are correctly inhibitory for \ssc{KICH}. Such granularity in heatmaps is helpful in understanding how the model arrives at a prediction and can prove to be useful for practitioners building the models as well as physicians using them.

\subsubsection{Model Debugging Using Additive MIL Heatmaps}
\begin{figure}[]
\centering
\includegraphics[scale=0.17]{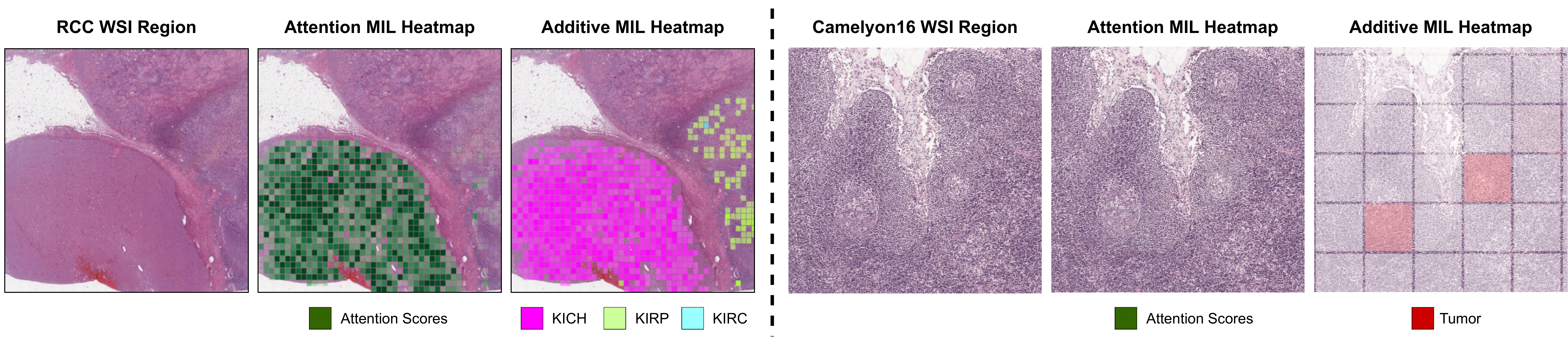}
\caption{Left $3$ columns show the case of model mis-predicting a \textcolor{LimeGreen}{\ssc{KIRP}} slide as \textcolor{Magenta}{\ssc{KICH}}. Attention heatmaps show a region of adrenal gland on the left being attended. Additive \ssc{MIL} heatmaps are able to exactly show how adrenal glands being rare, are being confused for \textcolor{Magenta}{\ssc{KICH}} regions even though the model correctly identifies the \textcolor{LimeGreen}{\ssc{KIRP}} regions on the right side. The right $3$ columns show a case from Camelyon16 where the model is mis-predicting a benign slide as malignant. The attention heatmap offers no information, however, Additive \ssc{MIL} heatmap highlights areas of germinal center as the source of this false positive prediction (in red). This pattern for false positive prediction is found in multiple other slides and can enable us to go from interpretation to debugging.}
\label{figure_debug_heatmaps}
\end{figure}

The ability of Additive \ssc{MIL} models to perfectly reflect model predictions at a patch-level can be useful in model debugging. Here, we show examples of the cases where Additive \ssc{MIL} heatmaps identify reasons for model failures during our experiments. Figure \ref{figure_debug_heatmaps} shows two such cases. These heatmaps not only provide interpretability to \ssc{MIL} models, but can also aid in validating specific hypothesis during model debugging.

\subsubsection{Limitations}
Since the interpretation of Additive \ssc{MIL} models is based on model predictions that are reformulated to be interpretable, the interpretability method is inherently coupled with the model. This is desirable since the heatmaps now exactly track model predictions, but this coupling also potentially limits the flexibility of the models and heatmaps. For example, since the patch contributions are tied to the model, one can only generate heatmaps with patches at the same resolution as what the model was trained on. Another limitation is the reduction of model expressivity introduced by the additive constraint. In this study, we did not find a practical example of this limitation, however, it may exist in other datasets.

\section{Conclusion \& Broader Impact}
We propose a simple reformulation of the popular Attention \ssc{MIL} setup for pathology that makes models intrinsically interpretable through an additive function. Our approach enables exact spatial credit assignment where the final prediction of the model can be attributed to individual contributions of each patch in a pathology slide. These models provide spatial interpretability without any loss of predictive performance and can be used for various applications like model debugging and highlighting regions-of-interest in a decision-support setting. This high fidelity interpretability will be critical in building trust for these models when deployed in medical decision-making.

\section*{Acknowledgements}
The authors would like to thank Francesco Rubbo for valuable feedback on the paper draft. The authors would also like to thank Aryan Pedawi and Harsha Pokkalla for providing early feedback on the idea.


\clearpage
\bibliography{ref}

\begin{thebibliography}{10}

\bibitem{aas2021explaining}
Kjersti Aas, Martin Jullum, and Anders L{\o}land.
\newblock Explaining individual predictions when features are dependent: More
  accurate approximations to shapley values.
\newblock {\em Artificial Intelligence}, 298:103502, 2021.

\bibitem{adebayo2018sanity}
Julius Adebayo, Justin Gilmer, Michael Muelly, Ian Goodfellow, Moritz Hardt,
  and Been Kim.
\newblock Sanity checks for saliency maps.
\newblock {\em Advances in neural information processing systems}, 31, 2018.

\bibitem{agarwal2021neural}
Rishabh Agarwal, Levi Melnick, Nicholas Frosst, Xuezhou Zhang, Ben Lengerich,
  Rich Caruana, and Geoffrey~E Hinton.
\newblock Neural additive models: Interpretable machine learning with neural
  nets.
\newblock {\em Advances in Neural Information Processing Systems}, 34, 2021.

\bibitem{anders2022finding}
Christopher~J Anders, Leander Weber, David Neumann, Wojciech Samek,
  Klaus-Robert M{\"u}ller, and Sebastian Lapuschkin.
\newblock Finding and removing clever hans: Using explanation methods to debug
  and improve deep models.
\newblock {\em Information Fusion}, 77:261--295, 2022.

\bibitem{atrey2019exploratory}
Akanksha Atrey, Kaleigh Clary, and David Jensen.
\newblock Exploratory not explanatory: Counterfactual analysis of saliency maps
  for deep reinforcement learning.
\newblock {\em arXiv preprint arXiv:1912.05743}, 2019.

\bibitem{bejnordi2017diagnostic}
Babak~Ehteshami Bejnordi, Mitko Veta, Paul~Johannes Van~Diest, Bram
  Van~Ginneken, Nico Karssemeijer, Geert Litjens, Jeroen~AWM Van Der~Laak,
  Meyke Hermsen, Quirine~F Manson, Maschenka Balkenhol, et~al.
\newblock Diagnostic assessment of deep learning algorithms for detection of
  lymph node metastases in women with breast cancer.
\newblock {\em Jama}, 318(22):2199--2210, 2017.

\bibitem{bosch2021machine}
Jaime Bosch, Chuhan Chung, Oscar~M Carrasco-Zevallos, Stephen~A Harrison,
  Manal~F Abdelmalek, Mitchell~L Shiffman, Don~C Rockey, Zahil Shanis, Dinkar
  Juyal, Harsha Pokkalla, et~al.
\newblock A machine learning approach to liver histological evaluation predicts
  clinically significant portal hypertension in nash cirrhosis.
\newblock {\em Hepatology}, 74(6):3146--3160, 2021.

\bibitem{bulten2021artificial}
Wouter Bulten, Maschenka Balkenhol, Jean-Jo{\"e}l~Awoumou Belinga, Am{\'e}rico
  Brilhante, Asl{\i} {\c{C}}ak{\i}r, Lars Egevad, Martin Eklund, Xavier
  Farr{\'e}, Katerina Geronatsiou, Vincent Molini{\'e}, et~al.
\newblock Artificial intelligence assistance significantly improves gleason
  grading of prostate biopsies by pathologists.
\newblock {\em Modern Pathology}, 34(3):660--671, 2021.

\bibitem{campanella2019clinical}
Gabriele Campanella, Matthew~G Hanna, Luke Geneslaw, Allen Miraflor, Vitor
  Werneck Krauss~Silva, Klaus~J Busam, Edi Brogi, Victor~E Reuter, David~S
  Klimstra, and Thomas~J Fuchs.
\newblock Clinical-grade computational pathology using weakly supervised deep
  learning on whole slide images.
\newblock {\em Nature medicine}, 25(8):1301--1309, 2019.

\bibitem{caruana2020intelligible}
Rich Caruana, Scott Lundberg, Marco~Tulio Ribeiro, Harsha Nori, and Samuel
  Jenkins.
\newblock Intelligible and explainable machine learning: best practices and
  practical challenges.
\newblock In {\em Proceedings of the 26th ACM SIGKDD International Conference
  on Knowledge Discovery \& Data Mining}, pages 3511--3512, 2020.

\bibitem{degrave2021ai}
Alex~J DeGrave, Joseph~D Janizek, and Su-In Lee.
\newblock Ai for radiographic covid-19 detection selects shortcuts over signal.
\newblock {\em Nature Machine Intelligence}, 3(7):610--619, 2021.

\bibitem{diao2021human}
James~A Diao, Jason~K Wang, Wan~Fung Chui, Victoria Mountain, Sai~Chowdary
  Gullapally, Ramprakash Srinivasan, Richard~N Mitchell, Benjamin Glass, Sara
  Hoffman, Sudha~K Rao, et~al.
\newblock Human-interpretable image features derived from densely mapped cancer
  pathology slides predict diverse molecular phenotypes.
\newblock {\em Nature communications}, 12(1):1--15, 2021.

\bibitem{echle2021deep}
Amelie Echle, Niklas~Timon Rindtorff, Titus~Josef Brinker, Tom Luedde,
  Alexander~Thomas Pearson, and Jakob~Nikolas Kather.
\newblock Deep learning in cancer pathology: a new generation of clinical
  biomarkers.
\newblock {\em British journal of cancer}, 124(4):686--696, 2021.

\bibitem{glass2021machine}
Benjamin Glass, Michel~Erminio Vandenberghe, Surya~Teja Chavali, Syed~Ashar
  Javed, Marlon Rebelatto, Shamira Sridharan, Hunter Elliott, Sudha Rao,
  Michael Montalto, Murray Resnick, et~al.
\newblock Machine learning models to quantify her2 for real-time tissue image
  analysis in prospective clinical trials., 2021.

\bibitem{hagele2020resolving}
Miriam H{\"a}gele, Philipp Seegerer, Sebastian Lapuschkin, Michael Bockmayr,
  Wojciech Samek, Frederick Klauschen, Klaus-Robert M{\"u}ller, and Alexander
  Binder.
\newblock Resolving challenges in deep learning-based analyses of
  histopathological images using explanation methods.
\newblock {\em Scientific reports}, 10(1):1--12, 2020.

\bibitem{hastie2017generalized}
Trevor~J Hastie and Robert~J Tibshirani.
\newblock {\em Generalized additive models}.
\newblock Routledge, 2017.

\bibitem{ilse2018attention}
Maximilian Ilse, Jakub Tomczak, and Max Welling.
\newblock Attention-based deep multiple instance learning.
\newblock In {\em International conference on machine learning}, pages
  2127--2136. PMLR, 2018.

\bibitem{javed2022rethinking}
Syed~Ashar Javed, Dinkar Juyal, Zahil Shanis, Shreya Chakraborty, Harsha
  Pokkalla, and Aaditya Prakash.
\newblock Rethinking machine learning model evaluation in pathology.
\newblock {\em arXiv preprint arXiv:2204.05205}, 2022.

\bibitem{kather2019deep}
Jakob~Nikolas Kather, Alexander~T Pearson, Niels Halama, Dirk J{\"a}ger,
  Jeremias Krause, Sven~H Loosen, Alexander Marx, Peter Boor, Frank Tacke,
  Ulf~Peter Neumann, et~al.
\newblock Deep learning can predict microsatellite instability directly from
  histology in gastrointestinal cancer.
\newblock {\em Nature medicine}, 25(7):1054--1056, 2019.

\bibitem{kindermans2019reliability}
Pieter-Jan Kindermans, Sara Hooker, Julius Adebayo, Maximilian Alber, Kristof~T
  Sch{\"u}tt, Sven D{\"a}hne, Dumitru Erhan, and Been Kim.
\newblock The (un) reliability of saliency methods.
\newblock In {\em Explainable AI: Interpreting, Explaining and Visualizing Deep
  Learning}, pages 267--280. Springer, 2019.

\bibitem{kingma2014adam}
Diederik~P Kingma and Jimmy Ba.
\newblock Adam: A method for stochastic optimization.
\newblock {\em arXiv preprint arXiv:1412.6980}, 2014.

\bibitem{leavitt2020towards}
Matthew~L Leavitt and Ari Morcos.
\newblock Towards falsifiable interpretability research.
\newblock {\em arXiv preprint arXiv:2010.12016}, 2020.

\bibitem{li2021dual}
Bin Li, Yin Li, and Kevin~W Eliceiri.
\newblock Dual-stream multiple instance learning network for whole slide image
  classification with self-supervised contrastive learning.
\newblock In {\em Proceedings of the IEEE/CVF Conference on Computer Vision and
  Pattern Recognition}, pages 14318--14328, 2021.

\bibitem{li2021multi}
Jiayun Li, Wenyuan Li, Anthony Sisk, Huihui Ye, W~Dean Wallace, William Speier,
  and Corey~W Arnold.
\newblock A multi-resolution model for histopathology image classification and
  localization with multiple instance learning.
\newblock {\em Computers in biology and medicine}, 131:104253, 2021.

\bibitem{linardatos2020explainable}
Pantelis Linardatos, Vasilis Papastefanopoulos, and Sotiris Kotsiantis.
\newblock Explainable ai: a review of machine learning interpretability
  methods.
\newblock {\em Entropy}, 23(1):18, 2020.

\bibitem{london2019artificial}
Alex~John London.
\newblock Artificial intelligence and black-box medical decisions: accuracy
  versus explainability.
\newblock {\em Hastings Center Report}, 49(1):15--21, 2019.

\bibitem{lu2021data}
Ming~Y Lu, Drew~FK Williamson, Tiffany~Y Chen, Richard~J Chen, Matteo Barbieri,
  and Faisal Mahmood.
\newblock Data-efficient and weakly supervised computational pathology on
  whole-slide images.
\newblock {\em Nature biomedical engineering}, 5(6):555--570, 2021.

\bibitem{lundberg2017unified}
Scott~M Lundberg and Su-In Lee.
\newblock A unified approach to interpreting model predictions.
\newblock {\em Advances in neural information processing systems}, 30, 2017.

\bibitem{ma2018shufflenet}
Ningning Ma, Xiangyu Zhang, Hai-Tao Zheng, and Jian Sun.
\newblock Shufflenet v2: Practical guidelines for efficient cnn architecture
  design.
\newblock In {\em Proceedings of the European conference on computer vision
  (ECCV)}, pages 116--131, 2018.

\bibitem{maron1997framework}
Oded Maron and Tom{\'a}s Lozano-P{\'e}rez.
\newblock A framework for multiple-instance learning.
\newblock {\em Advances in neural information processing systems}, 10, 1997.

\bibitem{messalas2019model}
Andreas Messalas, Yiannis Kanellopoulos, and Christos Makris.
\newblock Model-agnostic interpretability with shapley values.
\newblock In {\em 2019 10th International Conference on Information,
  Intelligence, Systems and Applications (IISA)}, pages 1--7. IEEE, 2019.

\bibitem{molnar2022general}
Christoph Molnar, Gunnar K{\"o}nig, Julia Herbinger, Timo Freiesleben, Susanne
  Dandl, Christian~A Scholbeck, Giuseppe Casalicchio, Moritz Grosse-Wentrup,
  and Bernd Bischl.
\newblock General pitfalls of model-agnostic interpretation methods for machine
  learning models.
\newblock In {\em International Workshop on Extending Explainable AI Beyond
  Deep Models and Classifiers}, pages 39--68. Springer, 2022.

\bibitem{pirovano2020improving}
Antoine Pirovano, Hippolyte Heuberger, Sylvain Berlemont, Sa{\"\i}d Ladjal, and
  Isabelle Bloch.
\newblock Improving interpretability for computer-aided diagnosis tools on
  whole slide imaging with multiple instance learning and gradient-based
  explanations.
\newblock In {\em Interpretable and Annotation-Efficient Learning for Medical
  Image Computing}, pages 43--53. Springer, 2020.

\bibitem{rudin2019stop}
Cynthia Rudin.
\newblock Stop explaining black box machine learning models for high stakes
  decisions and use interpretable models instead.
\newblock {\em Nature Machine Intelligence}, 1(5):206--215, 2019.

\bibitem{schmitt2021hidden}
Max Schmitt, Roman~Christoph Maron, Achim Hekler, Albrecht Stenzinger, Axel
  Hauschild, Michael Weichenthal, Markus Tiemann, Dieter Krahl, Heinz Kutzner,
  Jochen~Sven Utikal, et~al.
\newblock Hidden variables in deep learning digital pathology and their
  potential to cause batch effects: prediction model study.
\newblock {\em Journal of medical Internet research}, 23(2):e23436, 2021.

\bibitem{shao2021transmil}
Zhuchen Shao, Hao Bian, Yang Chen, Yifeng Wang, Jian Zhang, Xiangyang Ji,
  et~al.
\newblock Transmil: Transformer based correlated multiple instance learning for
  whole slide image classification.
\newblock {\em Advances in Neural Information Processing Systems}, 34, 2021.

\bibitem{shapley1953value}
Lloyd~S Shapley.
\newblock A value for n-person games, contributions to the theory of games, 2,
  307--317, 1953.

\bibitem{sundararajan2020many}
Mukund Sundararajan and Amir Najmi.
\newblock The many shapley values for model explanation.
\newblock In {\em International conference on machine learning}, pages
  9269--9278. PMLR, 2020.

\bibitem{tang2021data}
Siyi Tang, Amirata Ghorbani, Rikiya Yamashita, Sameer Rehman, Jared~A Dunnmon,
  James Zou, and Daniel~L Rubin.
\newblock Data valuation for medical imaging using shapley value and
  application to a large-scale chest x-ray dataset.
\newblock {\em Scientific reports}, 11(1):1--9, 2021.

\bibitem{trpkov2018benign}
Kiril Trpkov.
\newblock Benign mimics of prostatic adenocarcinoma.
\newblock {\em Modern Pathology}, 31(1):22--46, 2018.

\bibitem{van2022tractability}
Guy Van~den Broeck, Anton Lykov, Maximilian Schleich, and Dan Suciu.
\newblock On the tractability of shap explanations.
\newblock {\em Journal of Artificial Intelligence Research}, 74:851--886, 2022.

\bibitem{walk2009role}
Eric~E Walk.
\newblock The role of pathologists in the era of personalized medicine.
\newblock {\em Archives of pathology \& laboratory medicine}, 133(4):605--610,
  2009.

\bibitem{wang2016deep}
Dayong Wang, Aditya Khosla, Rishab Gargeya, Humayun Irshad, and Andrew~H Beck.
\newblock Deep learning for identifying metastatic breast cancer.
\newblock {\em arXiv preprint arXiv:1606.05718}, 2016.

\bibitem{wang2019machine}
Xiuying Wang, Dingqian Wang, Zhigang Yao, Bowen Xin, Bao Wang, Chuanjin Lan,
  Yejun Qin, Shangchen Xu, Dazhong He, and Yingchao Liu.
\newblock Machine learning models for multiparametric glioma grading with
  quantitative result interpretations.
\newblock {\em Frontiers in neuroscience}, page 1046, 2019.

\bibitem{weinstein2013cancer}
John~N Weinstein, Eric~A Collisson, Gordon~B Mills, Kenna~R Shaw, Brad~A
  Ozenberger, Kyle Ellrott, Ilya Shmulevich, Chris Sander, and Joshua~M Stuart.
\newblock The cancer genome atlas pan-cancer analysis project.
\newblock {\em Nature genetics}, 45(10):1113--1120, 2013.

\bibitem{wulczyn2020deep}
Ellery Wulczyn, David~F Steiner, Zhaoyang Xu, Apaar Sadhwani, Hongwu Wang,
  Isabelle Flament-Auvigne, Craig~H Mermel, Po-Hsuan~Cameron Chen, Yun Liu, and
  Martin~C Stumpe.
\newblock Deep learning-based survival prediction for multiple cancer types
  using histopathology images.
\newblock {\em PloS one}, 15(6):e0233678, 2020.

\bibitem{yao2020whole}
Jiawen Yao, Xinliang Zhu, Jitendra Jonnagaddala, Nicholas Hawkins, and Junzhou
  Huang.
\newblock Whole slide images based cancer survival prediction using attention
  guided deep multiple instance learning networks.
\newblock {\em Medical Image Analysis}, 65:101789, 2020.

\end{thebibliography}
\bibliographystyle{plain}
\clearpage

\appendix

\section{Appendix}
\label{shapley_proof}

\textbf{Theorem 1:} \textit{The marginal instance contribution from an Additive \ssc{MIL} model, $g(x_i)$ is proportional to the Shapley value of that instance, $\phi_i$.}
\begin{align*}
    g(x_i) \propto \phi_i(V, x) = \sum_{S\subseteq F \backslash i} \frac{|S!| (|F|-|S|-1|!)}{|F!|} V_{S \cup i}(x_{S \cup i}) - V_{S}(x_{S}) \numberthis \label{eqn_additive_shapley_appendix}
\end{align*}

\textit{Proof:} 
The interpretation for the value function $V_S$ is taken from \cite{lundberg2017unified} where it's defined as the expected value of the model given a specific input set $x^*_S$.
\begin{align*}
    V_S(x_S) = \mathbb{E}[g(x) | x_S = x^*_S]  \numberthis
\end{align*}
Since the conditional expectation is for the case where only the set $S$ is known, rewriting the equation in the form of integrals and breaking it down by set $S$ and its complement $\bar{S}$ gives:
\begin{align*}
    V_S(x_S) &= \int g(x) p(x_{\bar{S}} | x_S = x^*_S) dx_{\bar{S}}  \numberthis \label{eqn_v_1} \\
    &= \int \Big[ (\sum_{j \in S} g(x^*_j)) + (\sum_{j \in \bar{S}} g(x_j)) \Big] \: p(x_{\bar{S}} | x_S = x^*_S) dx_{\bar{S}}  \numberthis \label{eqn_v_2} \\
    &= \int \sum_{j \in S} g(x^*_j) \: p(x_{\bar{S}} | x_S = x^*_S) dx_{\bar{S}} + \int (\sum_{j \in \bar{S}} g(x_j) \: p(x_{\bar{S}} | x_S = x^*_S) dx_{\bar{S}}  \numberthis \label{eqn_v_3} \\
    &= \sum_{j \in S} g(x^*_j) \int p(x_{\bar{S}} | x_S = x^*_S) dx_{\bar{S}} + \sum_{j \in \bar{S}} \mathbb{E}[g(x_j)]  \numberthis \label{eqn_v_4} \\
    &= \sum_{j \in S} g(x^*_j) + \sum_{j \in \bar{S}} \mathbb{E}[g(x_j)] \numberthis \label{eqn_v_5}
\end{align*}
Equation \ref{eqn_v_2} uses the model definition from equation \ref{eqn_additive_mil_model} to express the function $g$ into its linearly additive components over all instances which are either in set $S$ or in $\bar{S}$. Similarly, we can write the value function when the $i^{th}$ index is included in $S$ by removing it from set $\bar{S}$ and adding it to $S$:
\begin{align*}
    V_{S \cup i}(x_{S \cup i}) &= V_S(x_S) + g(x^*_i) - \mathbb{E}[g(x_i)] \numberthis \label{eqn_v_6} \\
    V_{S \cup i}(x_{S \cup i}) - V_S(x_S) &= g(x^*_i) - \mathbb{E}[g(x_i)] \numberthis \label{eqn_v_7}
\end{align*}
Since the second term here is the expected value of the model output, we can put this back in equation \ref{eqn_additive_shapley_appendix} to get an equivalence between the Shapley value and the instance contribution from an Additive \ssc{MIL} model.
\begin{align*}
    \phi_i(V, x) \propto g(x_i)
\end{align*}

\clearpage


\begin{center}
\textbf{\LARGE Supplementary Material} \\
\end{center}

In the supplementary material, we provide additional heatmaps from \ssc{TCGA} and Camelyon16 datasets which compare attention and Additive \ssc{MIL} heatmaps on pathology whole-slide images. We also show results for an expert evaluation of the heatmaps' utility towards a clinical-grade tool.

\section{Heatmaps from TCGA-RCC Dataset}
\label{rcc}

\begin{figure}[p]
    \vspace*{-2cm}
    \makebox[\linewidth]{
        \includegraphics[width=1.15\linewidth,scale=0.02]{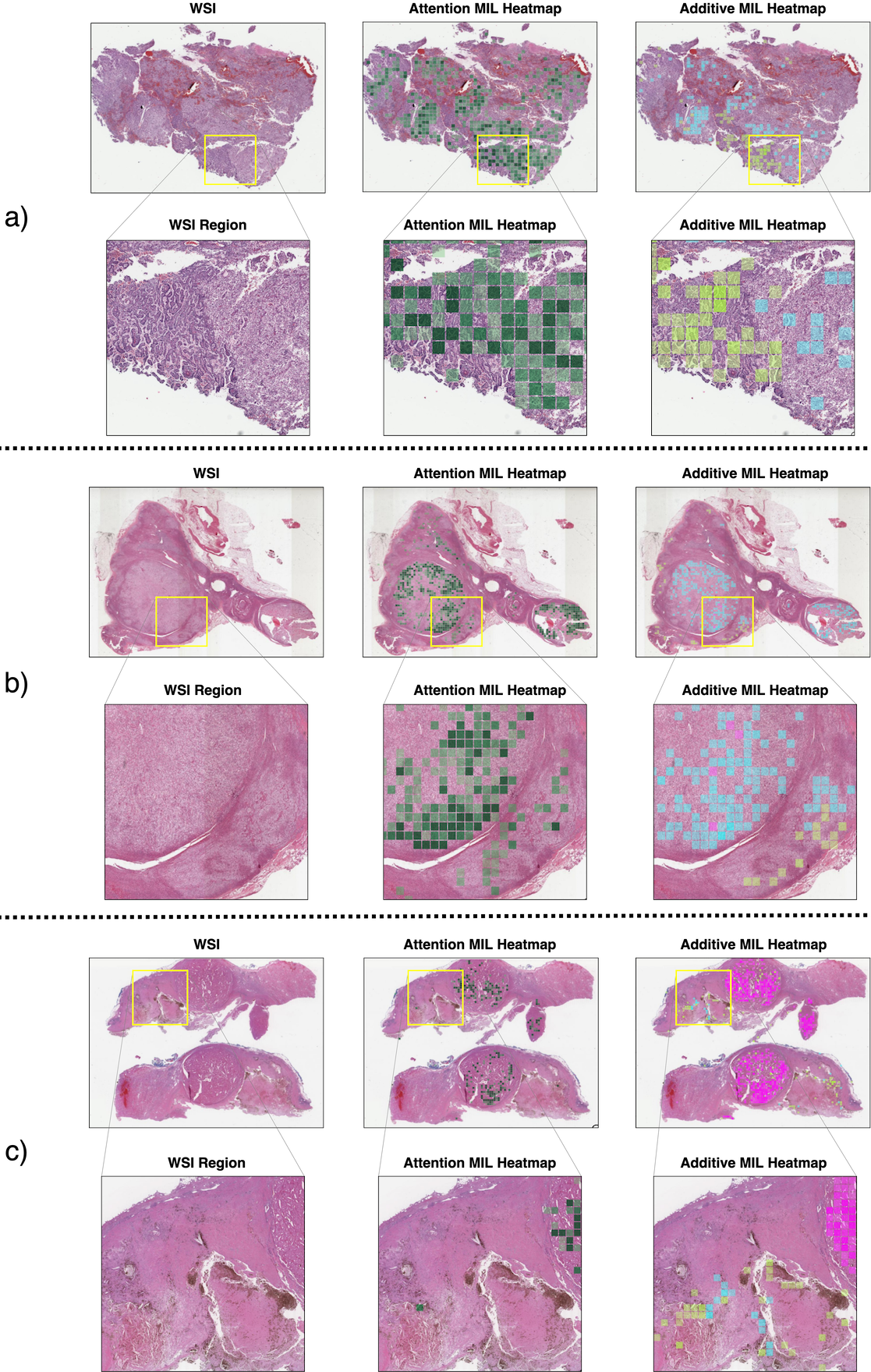}
    }
    \caption{Comparison of Additive \ssc{MIL} and attention heatmaps. Cyan patches denote \textcolor{Cyan}{\ssc{KIRC}}, lime green patches denote \textcolor{LimeGreen}{\ssc{KIRP}}, and pink patches denonte \textcolor{Magenta}{\ssc{KICH}}. Attention heatmaps are shown in green. Additive \ssc{MIL} heatmaps highlight regions different from attention heatmaps and offer more granularity in interpretation. See section \ref{rcc} for details about the shown cases.}
    \label{fig_rcc}
\end{figure}

Figure \ref{fig_rcc} shows three cases of Renal Cell Carcinoma (\ssc{RCC}). The \ssc{MIL} models are trained to predict the sub-type present in the slide, namely clear cell carcinoma (\ssc{KIRC}), papillary cell carcinoma (\ssc{KIRP}), \& chromophobe (\ssc{KICH}). In case (a), attention heatmap is shown to attend to regions predictive of both \ssc{KIRC} and \ssc{KIRP} whereas Additive \ssc{MIL} heatmap correctly identifies the presence of individual sub-types within the same slide. Similarly in (b), the slide shown is labeled as \ssc{KIRC}, however it contains areas with papillary structure as highlighted by the Additive \ssc{MIL} heatmap. Note that attention heatmap does not highlight these regions. In case (c), the slide is labeled as \ssc{KICH} and the model correctly predicts it. The attention heatmap highlights relevant \ssc{KICH} regions in pink. However, it misses showing patches contributing to the other two classes spuriously which are visible in the Additive \ssc{MIL} heatmap.

\section{Heatmaps from Camelyon16 Dataset}
\label{camelyon}
\begin{figure}[p]
    \vspace*{-2cm}
    \makebox[\linewidth]{
        \includegraphics[width=1.15\linewidth,scale=0.02]{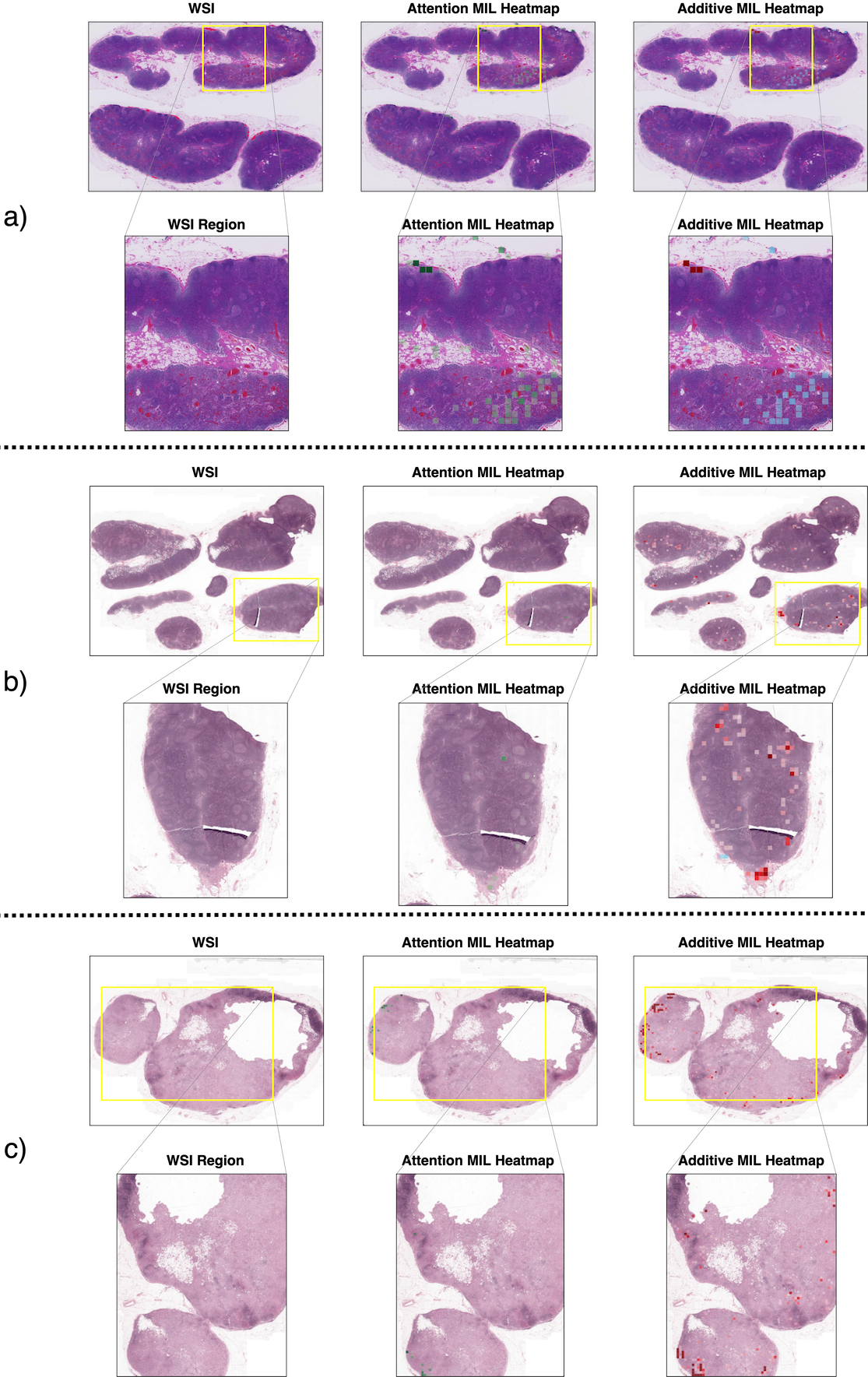}
    }
\caption{Comparison of Additive \ssc{MIL} and attention heatmaps. Red patches denote the class  \textcolor{Red}{\ssc{Malignant}} and blue patches denote the class \textcolor{Blue}{\ssc{Benign}}. Attention heatmaps are shown in green. Attention heatmaps do not distinguish between excitatory \& inhibitory patch contributions and often do not highlight false positive patches which are critical for model debugging. See section \ref{camelyon} for details about the shown cases.}
\label{fig_camelyon}
\end{figure}

Figure \ref{fig_camelyon} shows three cases of metastatic breast cancer from the Camelyon dataset. Case (a) shows a malignant slide where the model gives the correct prediction. Attention heatmap highlights certain regions for this prediction, but it's not clear whether the patch provides excitatory or inhibitory contribution for the malignant class. In contrast, Additive \ssc{MIL} heatmap shows that the patches in blue are inhibitory towards the predicted class. This highlights the a key limitation of attention heatmaps which show patch importance but not their predictive value towards or against a class. Case (b) is a Benign slide which is mis-predicted as Malignant. The attention heatmap does not highlight any regions, however the Additive MIL heatmap correctly identifies and localizes the false positive failure mode of the model. This makes Additive MIL models suitable for granular model debugging. Case (c) is a malignant slide correctly predicted by the model. The attention heatmap only localizes a single cancer focus on the left side of the slide even though the whole piece of tissue is malignant.
Additive MIL heatmap correctly identifies other cancer foci as well.

\section{Heatmaps from TCGA-NSCLC Dataset}
\label{nsclc}
\begin{figure}[p]
    \vspace*{-2cm}
    \makebox[\linewidth]{
        \includegraphics[width=1.15\linewidth,scale=0.02]{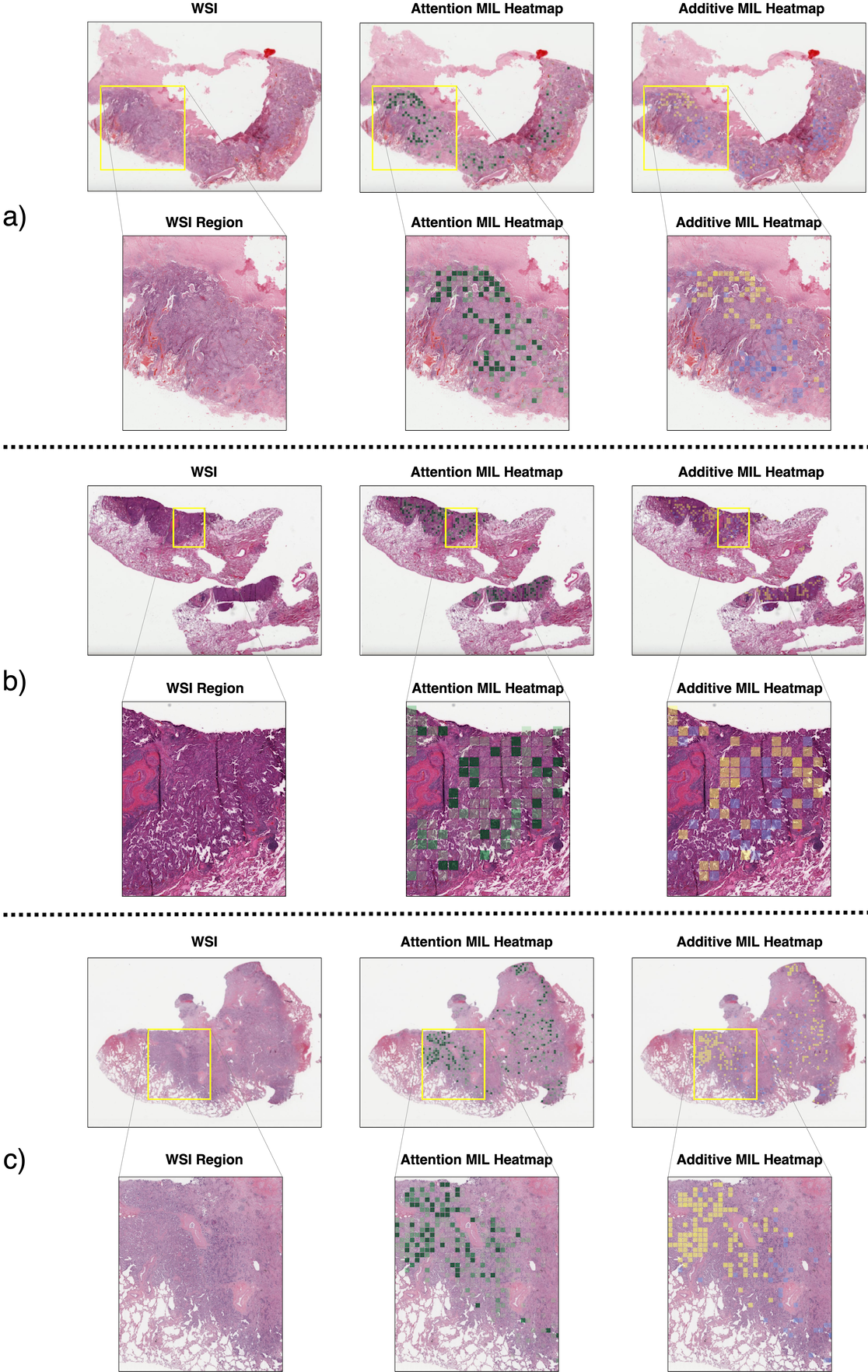}
    }
\caption{Comparison of Additive \ssc{MIL} and attention heatmaps. Yellow patches denote \textcolor{Goldenrod}{Adenocarcinoma} and blue patches denote \textcolor{Blue}{Squamous cell carcinoma}. Attention heatmaps are shown in green. They lack class-dependent patch attribution and often differ in their patch contribution values as compared to Additive MIL heatmaps. See section \ref{nsclc} for details about the shown cases.}
\label{fig_nsclc}
\end{figure}

Figure \ref{fig_nsclc} shows three cases of Non-Small Cell Lung Carcinoma (\ssc{NSCLC}). The \ssc{MIL} models are trained to predict the sub-type present in the slide, namely Adenocarinoma \& Squamous Cell Carcinoma. Additive \ssc{MIL} heatmap in (a) shows the model picking up regions predictive of both sub-types even though the model correctly predicts the slide to be Squamous Cell Carcinoma. This information is absent from attention heatmap. Note that in this case, the attention heatmap shows high importance for regions from both sub-types. In case (b) however, the most attended patches only correspond to Adenocarcinoma shown in yellow even though regions predictive of both sub-types exist. This ambiguous behavior of attention heatmap complicates interpretation. In (b), which is an Adenocarcinoma slide, Additive \ssc{MIL} heatmap shows the model being uncertain about the two classes and localizes that uncertainty to a specific region even though the final prediction is correct. In (c), we again see attention heatmap highlighting patches corresponding to both sub-types without distinguishing between them, while the Additive \ssc{MIL} heatmap clearly delineates the regions predictive of the two classes.

\section{Expert Evaluation of Additive Heatmaps \& Applicability in Decision Support Tool}
We conducted an expert evaluation of the heatmaps to assess their usefulness for highlighting regions of interest in Camelyon16 and \ssc{TCGA-RCC} slides using both Additive \ssc{MIL} and attention heatmaps. For Camelyon16, we selected a random sample of 50 slides from the test set with a 1:4 distribution of benign-to-malignant class. For \ssc{TCGA-RCC}, we randomly selected 39 slides with equal representation from all $3$ classes. A board-certified pathologist was asked to evaluate the heatmaps based on the following question:

\textit{"Which heatmap out of the two would you prefer to use in an AI+human decision-support setup for highlighting regions of interest before you give your diagnosis?"}

The results from the study are tabulated in Table \ref{tab_qualitative_study}. The scores for each heatmap are calculated by counting the number of times an expert pathologist would prefer one heatmap over the other. It clearly shows that Additive \ssc{MIL} heatmaps are almost always preferred over attention heatmaps. The main reason for this preference for \ssc{TCGA-RCC} was - "The Additive MIL heatmaps highlight patches for individual classes which can serve as visual reminder for pathologists to consider other differential diagnosis. "For Camelyon16, the pathologist feedback was - "Between Additive and Attention MIL, the former is preferred because the latter has more false positives and false negatives in all slides except one".

\begin{center}
\begin{table}[]
\caption{Expert evaluation of Additive and attention heatmaps for highlighting regions of interest in \ssc{TCGA-RCC} cancer sub-typing and Camelyon16 cancer identification task. The scores indicate the proportion of slides where a board-certified pathologist prefers a particular heatmap.}
\label{tab_qualitative_study}
\begin{tabular}{ |c||c|c| }
 \hline
 \textbf{Dataset} & \textbf{Attention Heatmap} & \textbf{Additive MIL Heatmap} \\ [2pt]
 \hline
 \hline
 TCGA-RCC & 6/39 & 33/39 \\[1pt]
 Camelyon16 & 1/50 & 49/50 \\[1pt] 
 \hline
\end{tabular}
\end{table}
\end{center}

\end{document}